\documentclass[runningheads]{llncs}
\usepackage[T1]{fontenc}
\usepackage{graphicx}
\usepackage{amsmath}
\usepackage{tikz}
\usepackage{pgfplots}
\usepackage{amsfonts}
\usepackage{subcaption}
\usepackage{todonotes}
\usepackage{array, makecell} %

\newcommand{\gat}{\mathrm{Ga}} 
\newcommand{\linear}{\mathrm{Lin}} 
\newcommand{\invivo}{{\em in vivo}} 
\newcommand{\exvivo}{{\em ex vivo}} 
\begin{document}

\title{SENDD: Sparse Efficient Neural Depth and Deformation for Tissue Tracking\thanks{This work was supported by Intuitive Surgical.}}
 \titlerunning{SENDD for Tissue Tracking}
\author{Adam Schmidt\inst{1}\orcidID{0000-0003-4769-4313} \and
Omid Mohareri\inst{2}\and
Simon DiMaio\inst{2}\and
Septimiu E. Salcudean\inst{1}}

\authorrunning{A. Schmidt et al.}
\institute{Department of Electrical and Computer Engineering, The University
of British Columbia, Vancouver, BC V6T 1Z4, Canada\\
\email{adamschmidt@ece.ubc.ca} \and
Advanced Research, Intuitive Surgical, Sunnyvale, CA 94086, USA}

%
\maketitle

\begin{abstract}
Deformable tracking and real-time estimation of 3D tissue motion is essential to enable automation and image guidance applications in robotically assisted surgery.
Our model, Sparse Efficient Neural Depth and Deformation (SENDD), extends prior 2D tracking work to estimate flow in 3D space.
SENDD introduces novel contributions of learned detection, and sparse per-point depth and 3D flow estimation, all with less than half a million parameters.
SENDD does this by using graph neural networks of sparse keypoint matches to estimate both depth and 3D flow anywhere.
We quantify and benchmark SENDD on a comprehensively labelled tissue dataset, and compare it to an equivalent 2D flow model.
SENDD performs comparably while enabling applications that 2D flow cannot.
SENDD can track points and estimate depth at 10fps on an NVIDIA RTX 4000 for 1280 tracked (query) points and its cost scales linearly with an increasing/decreasing number of points.
SENDD enables multiple downstream applications that require estimation of 3D motion in stereo endoscopy.
\end{abstract}

\section{Introduction}
Tracking of tissue and organs in surgical stereo endoscopy is essential to enable downstream tasks in image guidance~\cite{kaliaEvaluationMarkerlessIntraoperative2020}, surgical perception~\cite{giannarouVisionbasedDeformationRecovery2016,luSuperDeepSurgical2021}, motion compensation~\cite{richaRobust3DVisual2011}, and colonoscopy coverage estimation~\cite{zhangColDEDepthEstimation2021}.
Given the difficulty in creating labelled training data, we train an unsupervised model that can estimate motion for anything in the surgical field: tissue, gauze, clips, instruments.
Recent models for estimating deformation either use classical features and an underlying model (splines, embedded deformation, etc.~\cite{songMISSLAMRealTimeLargeScale2018,liSuPerSurgicalPerception2020}), or neural networks (eg. CNNs~\cite{teedRaftRecurrentAllpairs2020} or 2D graph neural networks (GNNs)~\cite{schmidtFastGraphRefinement2022}).
The issue with 2D methods is that downstream applications often require depth.
For example a correct physical 3D is needed location to enable augmented reality image guidance, motion compensation, and robotic automation (eg. suturing).
For our model, SENDD, we extend a sparse neural interpolation paradigm~\cite{schmidtFastGraphRefinement2022} to simultaneously perform depth estimation and 3D flow estimation.
This allows us to estimate depth and 3D flow all with one network rather than having to separately estimate dense depth maps.
With our approach, SENDD computes motion directly in 3D space, and parameterizes a 3D flow field that estimates the motion of any point in the field of view.

We design SENDD to use few parameters (low memory cost) and to scale with the number of points to be tracked (adaptive to different applications).
To avoid having to operate a 3D convolution over an entire volume, we use GNNs instead of CNNs.
This allows applications to tune how much computation to use by using more/fewer points.
SENDD is trained end-to-end.
This includes the detection, description, refinement, depth estimation, and 3D flow steps.
SENDD can perform frame-to-frame tracking and 3D scene flow estimation, but it could also be used as a more robust data association term for SLAM.
As will be shown in Section 2, unlike in prior work, our proposed approach combines feature detection, depth estimation and deformation modeling for scene flow all in one.
After providing relevant background for tissue tracking, we will describe SENDD, quantify it with a new IR-labelled tissue dataset, and finally demonstrate SENDD's efficiency.
The main novelties are that it is both \textbf{3D} and \textbf{Efficient} by: estimating scene flow anywhere in 3D space by using a GNN on salient points (\textbf{3D}), and reusing salient keypoints to calculate both sparse depth and flow at anywhere (\textbf{Efficient}).

\begin{figure}[t]
\centering
\includegraphics[width=\textwidth]{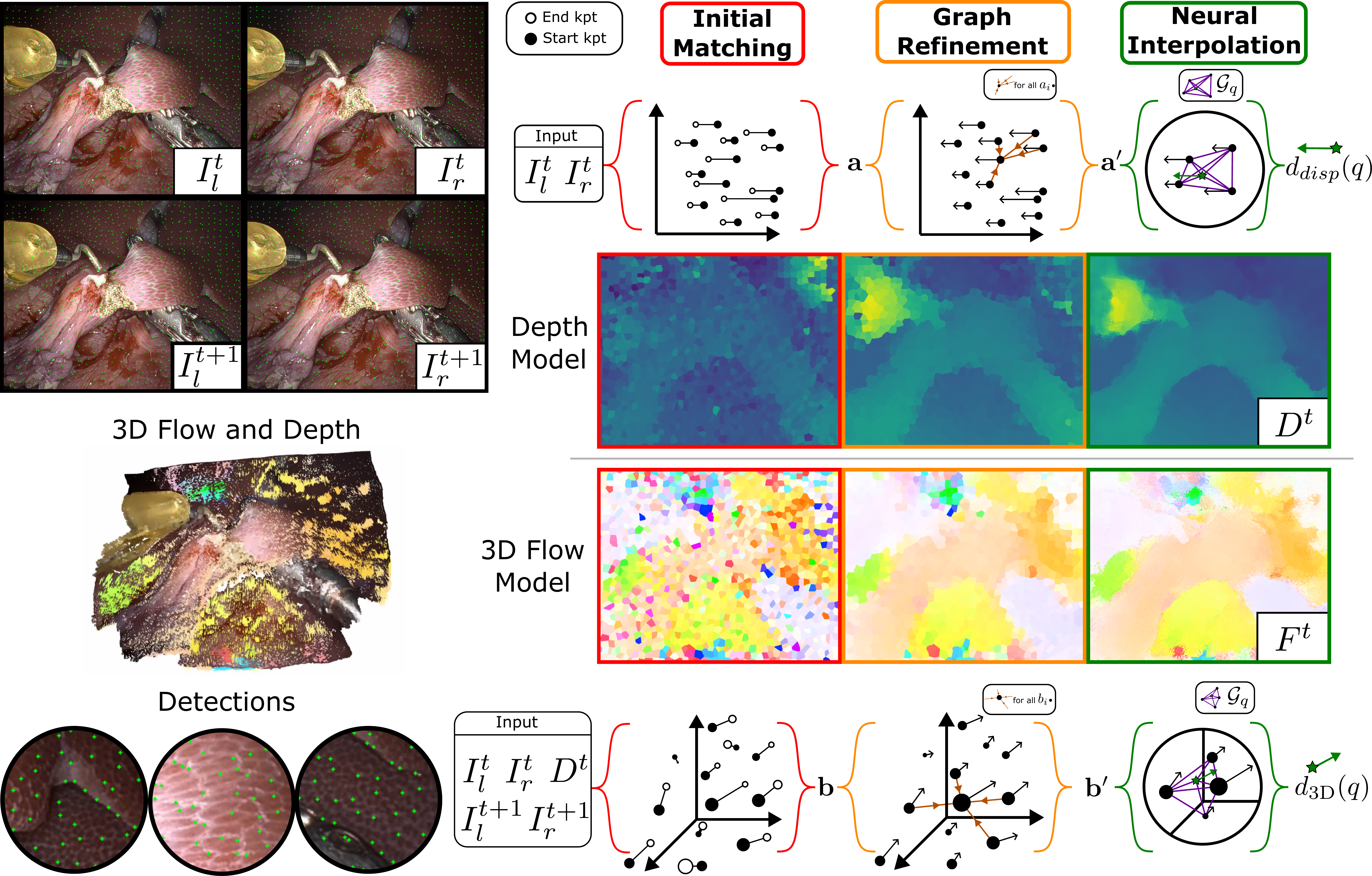}
	\caption{The SENDD model on a single query point, \(q\) (green star).
	From left to right are different steps in the refinement process for both depth (top) and flow (bottom). They both share feature points, but they use different weights for the graph refinement and interpolation steps. Both the depth and the flow network estimate motion of any point as a function of the nearest detected features using a GNN.
The depth of tracked (query) points is passed into the flow model to provide 3D position. \((\mathbf{a}, \mathbf{b})\) are learned  (depth, flow) features after nearest-neighbor keypoint matching, and \((\mathbf{a'}, \mathbf{b'})\) are node features after graph refinement.
 Only Neural Interpolation is repeated to track multiple queries.}\label{fig:flow}%
\end{figure}

\section{Background \& Related Work}

Different components are necessary to enable tissue tracking in surgery: feature detection, depth estimation, deformation modelling, and deformable Simultaneous Localization and Mapping (SLAM).
Our model acts as the feature detection, depth estimation, and deformation model for scene flow, all in one.

Recently, SuperPoint~\cite{detoneSuperPointSelfSupervisedInterest2018} features have been applied to endoscopy~\cite{barbedSuperPointFeaturesEndoscopy2022}.
These are trained using loss on image pairs that are warped with homographies.
In SENDD we use similar detections, but use a photometric reconstruction loss instead.
SuperGlue~\cite{sarlinSuperGlueLearningFeature2020} is a GNN method that can be used on top of SuperPoint to filter outliers, but it does not enable estimation of flow at non-keypoint locations, in addition to taking \({\sim} 270\)ms for 2048 keypoints.
Its GNN differs in that we use k-NN connected graph rather than a fully connected one.
In stereo depth estimation, BDIS~\cite{songBDISBayesianDense2023} introduces efficient improvements on classical methods, running in \({\sim}{70}\,\textrm{ms}\) for images of size \((1280, 720)\).
For flow estimation, there are the CNN-based RAFT~\cite{teedRaftRecurrentAllpairs2020}, and RAFT3D~\cite{teedRAFT3DSceneFlow2021} (45M params \({\sim} 386\)ms) which downsample by 8x and require computation over full images.
KINFlow~\cite{schmidtFastGraphRefinement2022} estimates motion using a GNN, but only in 2D.
For SLAM in endoscopy, MIS-SLAM~\cite{songMISSLAMRealTimeLargeScale2018} uses classical descriptors and models for deformation.
More recent work still does this, mainly due to the high cost of using CNNs when only a few points are actually salient.
Specifically, Endo-Depth-and-Motion~\cite{recasensEndoDepthandMotionReconstructionTracking2021} performs SLAM in rigid scenes.
DefSLAM and SD-DefSLAM~\cite{lamarcaDefSLAMTrackingMapping2021,gomez-rodriguezSDDefSLAMSemiDirectMonocular2021} both use meshes along with ORB features or Lucas-Kanade optical flow, respectively, for data association.
Lamarca et al.~\cite{lamarcaDirectSparseDeformable2022} track surfels using photometric error, but they do not have an underlying interpolation model for estimating motion between surfels.
SuPer~\cite{liSuPerSurgicalPerception2020}, and its extensions~\cite{linSemanticSuPerSemanticawareSurgical2023,luSuperDeepSurgical2021} use classical embedded deformation for motion modelling (\({\sim} 500\) ms/frame).
Finally, for full scene and deformation estimation, NERF~\cite{mildenhallNeRFRepresentingScenes2021} has recently been applied to endoscopy~\cite{wangNeuralRenderingStereo2022a}, but requires per-scene training and computationally expensive.
With SENDD, we fill the gap between classical and learned models by providing a flexible and efficient deformation model that could be integrated into SuPer~\cite{liSuPerSurgicalPerception2020}, SLAM, or used for short-term tracking.


\begin{figure}[tbp]
\hspace*{\fill}%
\begin{subfigure}[t]{.39\textwidth}%
\centering
\includegraphics[width=\textwidth]{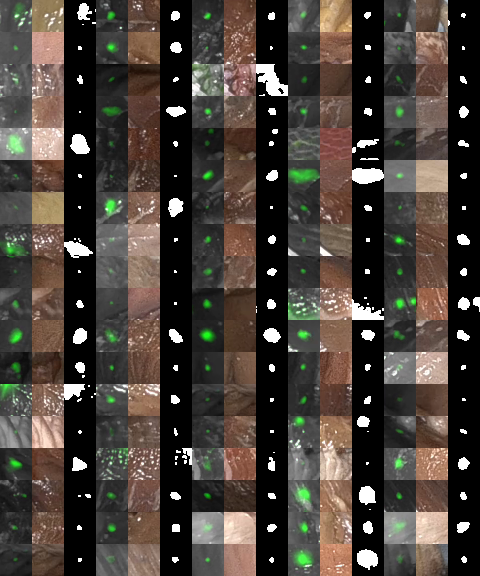}
\end{subfigure}%
\hfill
\begin{subfigure}[t]{.585\textwidth}
\centering
\includegraphics[width=\textwidth]{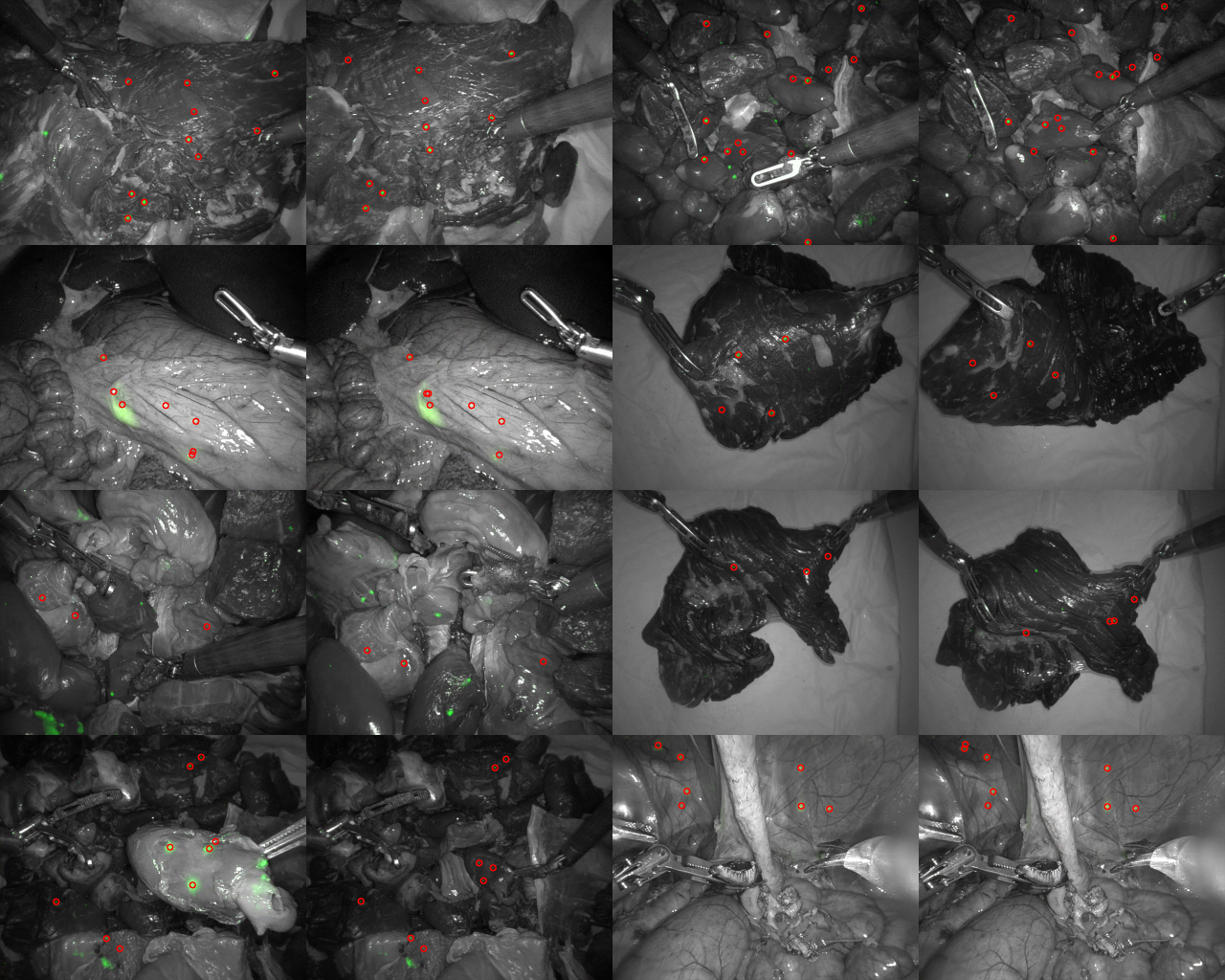}%
\end{subfigure}%
\hspace*{\fill}%
\caption{Test dataset images. Left: patch triplets of IR ims., visible light ims., and ground truth segmentations. Right: IR ims. from the start and end of clips with circles around labelled segment centers. Dataset statistics  (\invivo{} 266 min./\exvivo{} 134 min.) with (182/1139) clips and (944/9596) segments.} \label{patches}
\end{figure}

\section{Methods}
The SENDD model is described in Fig.~\ref{fig:flow}.
SENDD improves on prior sparse interpolation methods by designing a solution that: \textbf{learns a detector as part of the network, uses detected points to evaluate depth in a sparse manner, and uses a sparse interpolation paradigm to evaluate motion in 3D instead of 2D}.
SENDD consists of two key parts, with the first being the 1D stereo interpolation (Depth), and the second being the 3D flow estimation (Deformation).
Both use GNNs to estimate query disparity/motion using a coordinate-based Multi Layer Perceptron (MLP) (aka. implicit functions~\cite{tancikFourierFeaturesLet2020}) of keypoints neighboring it.
Before either interpolation step, SENDD detects and matches keypoints.
After detailing the detection step, we will explain the 3D flow and disparity models.
For the figures in this paper, we densely sample query points on a grid to show dense visualizations, but in practice the query points can be user or application defined.
Please see the supplementary video for examples of SENDD tracking tissue points.

\textbf{Learned Detector:}
Like SuperPoint~\cite{detoneSuperPointSelfSupervisedInterest2018}, we detect points across an image, but we use fewer layers and estimate a single location for each \((32, 32)\) region (instead of \((8, 8)\)).
Our images \((I^t_l, I^{t}_r)\) are of size \((1280, 1024)\), so we have \(N=1280\) detected points per frame. 
Unlike SuperPoint or others in the surgical space~\cite{barbedSuperPointFeaturesEndoscopy2022}, we do not rely on data augmentation for training, and instead allow the downstream loss metric to optimize detections to best reduce the final photometric reconstruction error.
Thus the detections are directly trained on real-world deformations.
Since we have a sparse regime, we use a softmax dot product score in training, and use \(\mathrm{max}\) in inference.
See Fig.~\ref{fig:flow} for examples of detections.

\textbf{3D Flow Network:}
SENDD estimates the 3D flow \(d_{\mathrm{3D}}(q) \in \mathbb{R}^3\) for each query \(q\in \mathbb{R}^2\) using two images \((I^t_l, I^{t+1}_l)\), and their depth maps  \((D^t, D^{t+1})\).
For a means to query depth at arbitrary points, see the section on sparse depth estimation.
To obtain initial matches, we match the detected points in 2D from frame \(I^t_l\) to \(I^{t+1}_l\).
We use ReTRo keypoints~\cite{schmidtRealTimeRotatedConvolutional2021} as descriptors by training the ReTRo network paired with SENDD's learned detector.
We do this to remain as lightweight as possible.
Matches further than 256 pixels away are filtered out.
This results in \(N\) pairs of points in 2D with positions \(p^{2D}_i, p'^{2D}_{i}\) and feature descriptors \(f_i, f_i',  \{i \in {1,\ldots, N}\}, f_i \in \mathbb{R}^c\).
These pairs are nodes in our graph, \(\mathcal{G}\).
Given the set of preliminary matches, we can then get their 3D positions by using our sparse depth interpolation and backprojecting using the camera parameters at each point \(p^{2D}_i, p'^{2D}_{i}\rightarrow p^{3D}_i, p'^{3D}_i\).
The graph, \(\mathcal{G}\), is defined with edges connecting each node to its k-nearest neighbors (k-NN) in 3D, with an optional dilation to enable a wider field of view at low cost.
The positions and features of each correspondence are combined to act as features in this graph.
For each node (detected point in image \(I^t_l\)), its feature is: \(
    b_i = \phi_3(p^{3D}_i) + \phi_3(p'^{3D}_i) + \gamma_d(f_i) + \gamma_e(f'_i) + \phi_1(||f_i - f'_i||_2) + \phi_1(||p_i-p'_i||_2)\).
\(\phi_{dim}: \mathbb{R}^{dim} \to \mathbb{R}^c\), denotes a positional encoding layer~\cite{tancikFourierFeaturesLet2020}, and \(\gamma_*\) are linear layers followed by a ReLU, where different subscripts denote different weights.
\(b_i\) are used as features in our graph attention network, with the bold version defined as the set of all node features \(\mathbf{b} = \left\{b_i | i \in N\right\}\).
By using a graph-attention neural network (GNN), as described  in~\cite{schmidtFastGraphRefinement2022}, we can refine this graph in 3D to estimate refined offsets and higher level local-neighborhood features.
\(\mathbf{b}' = \gat{}\left(\gat\left(\gat\left(\gat\left(\mathbf{b}, \mathcal{G}\right)\right)\right)\right)\), is the final set of node features that are functions of their neighborhood, \(\mathcal{N}\), in the graph.
\(\gat\) is the graph attention operation.
In practice, for each layer we use dilations of \([1, 8, 8, 1]\), and \(k=4\).
The prior steps only need to run once per image pair.
The motion \(d_{\mathrm{3D}}(q)\) of each query point \(q\) is a function of the nearby nodes in 3D, with \(\mathcal{G}_q\) denoting a k-clique graph of the query point and its \(k-1\) nearest nodes;
\(d_{\mathrm{3D}}(q) \in \mathbb{R}^3 = \linear_{3D}\left(\gat\left(\{q\} \mathbin\Vert \{\textbf{b}'_\mathcal{N}\}, \mathcal{G}_q\right)\right)\).
\(\linear_{3D}\) is a linear layer that converts from \(c\) channels to 3.

\textbf{Sparse depth interpolation:}
Instead of running a depth CNN in parallel, we estimate disparity sparsely as needed by using the same feature points with another lightweight GNN.
We do this as we found CNNs (eg. GANet~\cite{zhangGANetGuidedAggregation2019}) too  expensive in terms of training and inference time.
We adapt the 3D GNN interpolation methodology to estimate 1D flow along epipolar lines in the same way that we modified a 2D sparse flow model to work in 3D.
First, matches are found along epipolar lines between left and right images \((I^t_r, I^t_l)\).
Then matches are refined and query points are interpolated using a GNN.
\(\mathbf{a}, \mathbf{a'}\) are the 1D equivalents of \(\mathbf{b}, \mathbf{b'}\) from the 3D flow network.
For the refined node features, \(\mathbf{a}' \in \mathbb{R}^1 = \gat\left(\gat\left(\gat\left(\gat\left(\textbf{a}, \mathcal{G}\right)\right)\right)\right)\), and the final disparity estimate, \(d_{\mathrm{disp}}(q) \in \mathbb{R}^1 = \linear_{1D}\left(\gat\left(\left\{q\right\} \mathbin\Vert \left\{\mathbf{a}'_\mathcal{N}\right\}, \mathcal{G}_q\right)\right)\).
\(\linear_{1D}\) is a linear layer that converts from \(c\) channels to 1.
This can be seen like a neural version of the classic libELAS~\cite{geigerEfficientLargeScaleStereo2011}, where libELAS uses sparse support points on a regular grid along Sobel edge features to match points within a search space defined by the Delauney triangulation of support point matches.

\textbf{Loss:}
We train SENDD using loss on the warped stereo and flow images:

{\allowdisplaybreaks
\begin{align}
\mathcal{L}_{p}(A,B) &= \alpha \frac{1 - SSIM(A, B)}{2} + (1 - \alpha)\|A - B\|_1.\\
\mathcal{L}_{s}(V) &= \frac{1}{n} \sum \exp\left( -\frac{\beta}{3}\sum_c\left| \frac{\partial I}{\partial x}\right|\right) \left|\frac{\partial V}{\partial x}\right| +  \exp\left(-\frac{\beta}{3}\sum_c\left| \frac{\partial I}{\partial y}\right|\right) \left|\frac{\partial V}{\partial y}\right| \\
\mathcal{L}_{d} &= \left |D^{t\rightarrow t-1} - D^{t-1}\right | \\
\mathcal{L}_{total} &=  \mathcal{L}_{p}(I^{l}, I^{r\rightarrow l})+ \mathcal{L}_{p}(I^{t}, I^{t+1 \rightarrow t}) + \lambda_{F} \mathcal{L}_{s}(F) + \lambda_{D} \mathcal{L}_{s}(D) + \lambda_d \mathcal{L}_{d}
\end{align}
 \(\mathcal{L}_{p}(A,B)\) is a photometric loss function of input images \((A, B)\) that is used for both the warped depth pairs (\(\mathcal{L}_{p}(I^{l}, I^{r\rightarrow l})\)) and warped flow pairs (\(\mathcal{L}_{p}(I^{t}, I^{t+1 \rightarrow t})\)),
\(\mathcal{L}_{s}(V)\) is a smoothness loss on a flow field \(V\), and \(c\) denotes image color channels~\cite{jonschkowskiWhatMattersUnsupervised2020}. These loss functions are used for both the warped depth \((I^{l}, I^{r\rightarrow l})\) and the flow reconstruction pairs \((I^{t}, I^{t+1 \rightarrow t})\).
\(F\) and \(D\) are densely queried images of flow and depth.
We add a depth matching loss, \(\mathcal{L}_{d}\) which encourages the warped depth map to be close to the estimated depth map.
We set \(\alpha = 0.85, \beta = 150, \lambda_d = 0.001, \lambda_{F} = 0.01, \lambda_{D} = 1.0\), using values similar to~\cite{jonschkowskiWhatMattersUnsupervised2020}, and weighting depth matching lightly as a guiding term.
}

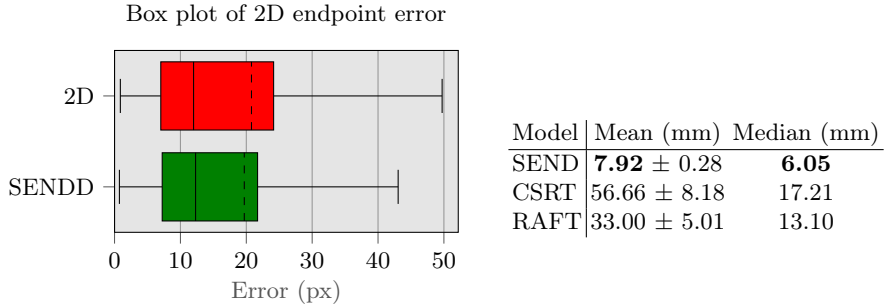
\begin{figure}[tbp]
\centering
\begin{subfigure}[t]{.56\textwidth}%
\centering
\begin{tikzpicture}

\definecolor{dimgray85}{RGB}{85,85,85}
\definecolor{gainsboro229}{RGB}{229,229,229}
\definecolor{dimgray200}{RGB}{150,150,150}
\definecolor{green}{RGB}{0,128,0}
\definecolor{mediumpurple152142213}{RGB}{152,142,213}
\definecolor{steelblue52138189}{RGB}{52,138,189}

\begin{axis}[
axis background/.style={fill=gainsboro229},
	title={Box plot of 2D endpoint error},
axis line style={black},
height=4cm,
tick align=outside,
tick pos=left,
width=0.9\textwidth,
x grid style={dimgray200},
	xlabel=\textcolor{dimgray85}{Error (px)},
xtick distance={10},
xmajorgrids,
xmin=0, xmax=52.1795951304625,
xtick style={color=dimgray85},
y grid style={dimgray85},
ymin=0.5, ymax=2.5,
ytick style={color=dimgray85},
ytick={1,2},
yticklabels={SENDD, 2D},
]
\addplot [black]
table {%
7.2327127871305 1
0.735544948048633 1
};
\addplot [black]
table {%
21.7133728042572 1
43.0540731694142 1
};
\addplot [black]
table {%
0.735544948048633 0.8125
0.735544948048633 1.1875
};
\addplot [black]
table {%
43.0540731694142 0.8125
43.0540731694142 1.1875
};
\addplot [black]
table {%
7.01956682231841 2
0.878890710046842 2
};
\addplot [black]
table {%
24.1490285615702 2
49.7298784551094 2
};
\addplot [black]
table {%
0.878890710046842 1.8125
0.878890710046842 2.1875
};
\addplot [black]
table {%
49.7298784551094 1.8125
49.7298784551094 2.1875
};
\path [draw=black, fill=green]
(axis cs:7.2327127871305,0.625)
--(axis cs:7.2327127871305,1.375)
--(axis cs:21.7133728042572,1.375)
--(axis cs:21.7133728042572,0.625)
--(axis cs:7.2327127871305,0.625)
--cycle;
\path [draw=black, fill=red]
(axis cs:7.01956682231841,1.625)
--(axis cs:7.01956682231841,2.375)
--(axis cs:24.1490285615702,2.375)
--(axis cs:24.1490285615702,1.625)
--(axis cs:7.01956682231841,1.625)
--cycle;
\addplot [black]
table {%
12.3008116853198 0.625
12.3008116853198 1.375
};
\addplot [black, dashed]
table {%
19.6948189846117 0.625
19.6948189846117 1.375
};
\addplot [black]
table {%
11.9973081896116 1.625
11.9973081896116 2.375
};
\addplot [black, dashed]
table {%
20.7881132858983 1.625
20.7881132858983 2.375
};
\end{axis}

\end{tikzpicture}
\end{subfigure}%
\hfill
	\begin{subfigure}[b]{.43\textwidth}
\centering
	\begin{tabular}[b]{l|cc}
		Model	& Mean (mm)			& Median (mm)\\
	\hline
	SEND	& \textbf{7.92} \(\pm\) 0.28	& \textbf{6.05}\\
	CSRT	& 56.66 \(\pm\) 8.18	& 17.21\\
	RAFT	& 33.00 \(\pm\) 5.01	& 13.10\\
\end{tabular}
		\vspace{1cm}

\label{tab:datastats}
\end{subfigure}%
	\caption{Endpoint error over the full dataset. Left: 2D endpoint error in a boxplot. Mean is the dotted line. Right: 3D endpoint error compared to using RAFT or CSRT to track in both left and right frames. Standard error of the mean is denoted with \(\pm\).} \label{fig:endpointerror}
\end{figure}

\section{Experiments}

We train SENDD with a set of rectified stereo videos from porcine clinical labs collected with a da Vinci Xi surgical system.
We randomly select images and skip between \((1,45)\) frames for each training pair.
We train using PyTorch with a batch size of 2 for 100,000 steps using a one-cycle learning rate schedule with maxlr=1e-4, minlr=4e-6.
We use 64 channels for all graph attention operations.

\textbf{Dataset:}
In order to quantify our results, we generate a new dataset.
The primary motivation for this is to have metrics that are not dependent on visible markers or require human labelling (which can bias to salient points).
Some other datasets have points that are hand labelled in software~\cite{liSuPerSurgicalPerception2020,cartuchoSurgTSoftTissueTracking2023} or use visible markers~\cite{linSemanticSuPerSemanticawareSurgical2023}. 
Others rely on RMSE of depth maps as a proxy for error, but this does not account for tissue sliding.
Our dataset uses tattooed points that flouresce under infrared (IR) light by using ICG dye.
For each ground truth clip, we capture an IR frame at the start, record a deformation action, and then capture another IR frame.
The start and end IR segments are then used as ground truth for quantification.
Points that are not visible at both the start and finish are then manually removed.
See Fig.~\ref{patches} for examples of data points at start and end frames.
The dataset includes multiple different tissue types in \exvivo{}, including: stomach, kidney, liver, intestine, tongue, heart, chicken breast, pork chop, etc.
Additionally the dataset includes recordings from four different \invivo{} porcine labs.
This dataset will be publicly released separately before MICCAI.
No clips from this labelled dataset are used in training.

\textbf{Quantification:} Instead of using Intersection Over Union (IOU) which would fail on small segments (eg. \(IOU=0\) for a one pixel measurement that is one pixel off), we use endpoint error and chamfer distance between segments to quantify SENDD's performance.
We compute endpoint error between the centers of each segmentation region.
We use chamfer distance as well because it allows us to see error for regions that are larger or non-circular.
Chamfer distance provides a metric that has an lower bound on error, in that if true error is zero then this will also be zero.

\begin{figure}[tbp]
\centering
\input{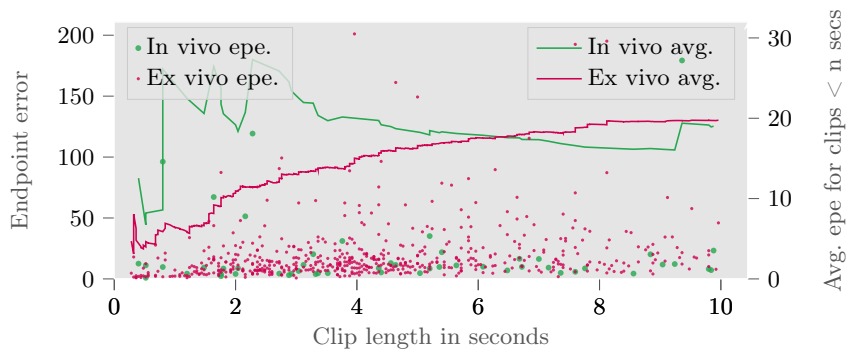}
\caption{{\em In vivo} and \exvivo{} errors of SENDD over the dataset.} \label{fig:invivovsexvivo}
\end{figure}

\textbf{Experiments:} The primary motivation for creating a 3D model (SENDD) instead of 2D is that it enables applications that require understanding of motion in 3D (automation, image guidance models).
That said, we still compare to the 2D version to compare performance in pixel space.
To quantify performance on clips, since these models only estimate motion for frame pairs, we run SENDD for each frame pair over \(n\) frames in a clip, where \(n\) is the clip length. No relocalization or drift prevention is incorporated (the supplementary material shows endpoint error on strides other than one).
For all experiments we select clips with length less than 10 seconds, as we are looking to quantify short term tracking methods.
The 2D model has the exact same parameters and model structure as SENDD (3D), except it does not do any depth map calculation, and only runs in image space.
First, we compare SENDD to the equivalent 2D model.
We do this with endpoint error as seen in Fig.~\ref{fig:endpointerror}, and show that the 3D method outperforms 2D over the full dataset.
Additionally, we compare to baselines of CSRT~\cite{lukezicDiscriminativeCorrelationFilter2017} which is a high performer in the SurgT challenge~\cite{cartuchoSurgTSoftTissueTracking2023} and RAFT~\cite{teedRaftRecurrentAllpairs2020}.
To track with CSRT and RAFT, we track in each left and right frame and backproject as done in SurgT~\cite{cartuchoSurgTSoftTissueTracking2023}.
SENDD outperforms these methods used off-the-shelf; see Fig.~\ref{fig:endpointerror}.
Performance of the RAFT and CSRT methods could likely be improved with a drift correction, or synchronization between left and right to prevent depth from becoming more erroneous.
Then we compare chamfer distance. 
The 3D method also outperforms the 2D method, on 64 randomly selected clips, with a total average error of 19.18 pixels vs 20.88 pixels.
The reasons for the SENDD model being close in performance to the equivalent 2D model could be that the 3D method uses the same amount of channels to detect and describe the same features that will be used for both depth and flow.
Additionally, lens smudges or specularities can corrupt the depth map, leading to errors in the 3D model that the purely photometric 2D model might not encounter.
In 3D all it takes is for one camera artifact to in either image to obscure the depth map, and the resulting flow.
The 2D method decreases the likelihood of this happening as it only uses one image.
Finally, we compare performance in terms of endpoint error of our 3D model on \invivo{} vs \exvivo{} labelled data.
As is shown in Fig.~\ref{fig:invivovsexvivo}, the \invivo{} experiments have a more monotonic performance decrease relative to clip length.
Actions in the \exvivo{} dataset were performed solely to evaluate tracking performance, while the \invivo{} data was collected alongside training while performing standard surgical procedures (eg. cholecystectomy).
Thus the \invivo{} scenes can have more complicated occlusions or artifacts that SENDD does not account for, even though it is also trained on \invivo{} data.

\begin{table}[tbp]
\caption{Times in milliseconds for using SENDD with 1280 detected keypoints to track 1280 query points. Times are averaged over 64 runs, with standard error in the second row. Streaming total is calculated by subtracting \((\frac{1}{2}(\mathrm{col1} + \mathrm{col2}) + \mathrm{col4})\) from the stereo and flow total. This is the time that it would take if reusing features, as is enabled in a streaming manner (10fps).}
\begin{center}
\begin{tabular}{cccccc|c}
\makecell{Detection \&\\ Description (x4)} & \makecell{Auxiliary\\Buffer (x4)}& \makecell{3. Stereo \\Refinement} & \makecell{Stereo \\Total} & \makecell{Flow\\ Refinement} & \makecell{Stereo \&\\ Flow Total} & \makecell{Streaming\\Total}\\
\hline
50.1  & 1.8 & 12.9 & 21.4 & 32.3  & 145.4 & 98.1\\
\(\pm\) 0.4 & \(\pm\) 0.02 & \(\pm\) 0.6 & \(\pm\) 0.8 & \(\pm\) 0.9 & \(\pm\) 2.1 & \\
\end{tabular}
\end{center}
\label{tab:datastats}
\end{table}

\textbf{Benchmarking and Model Size: }
SENDD has only \(366{,}195\) parameters, compared to other models which estimate just flow (RAFT-s~\cite{teedRaftRecurrentAllpairs2020} with 1.0M params.) or depth (GANet~\cite{zhangGANetGuidedAggregation2019} with 0.5M params.) using CNNs.
We benchmark SENDD on a NVIDIA Quadro RTX 4000, with a batch size of one, 1280 query points, and 1280 control points (keypoints).
As seen in Table~\ref{tab:datastats}, the stereo estimation for each frame takes 21.4 ms, and total time for the whole network to estimate stereo (at both t and t+1) and flow is 145.4 ms.
When estimating flow for an image pair pair, we need to calculate two sets of keypoint features, but for a video, we can reuse the features from the previous frame instead of recalculating the full pair each time.
Results with reuse are exactly the same, only timing differs.
Subtracting out the time for operations that do not need to be repeated leaves us with a streaming time of 97.3 ms, with a frame rate of 10fps for the entire flow and depth model.
This can be further improved by using spatial data structures for nearest neighbor lookup or PyTorch optimizations that have not been enabled (eg. float16).
The number of salient (or query) points can also be changed to adjust refinement (or neural interpolation) time.

\section{Conclusion}
SENDD is a flexible model for estimating deformation in 3D.
A limitation of SENDD is that it is unable to cope with occlusion or relocalization, and like all methods is vulnerable to drift over long periods.
These could be amended by integrating a SLAM system. 
We demonstrate that SENDD performs better than the equivalent sparse 2D model while additionally enabling parameterization of deformation in 3D space, learned detection, and depth estimation.
SENDD enables real-time applications that can rely on tissue tracking in surgery, such as long term tracking, or deformation estimation.

\bibliographystyle{splncs04}
\bibliography{SENDD}

\end{document}


\title{Supplementary Material for SENDD: Sparse Efficient Neural Depth and Deformation for Tissue Tracking}
 \titlerunning{Supplementary Material for SENDD}
\author{}
\institute{}

%
\maketitle
\begin{figure}[htbp]
\centering
\includegraphics[width=\textwidth]{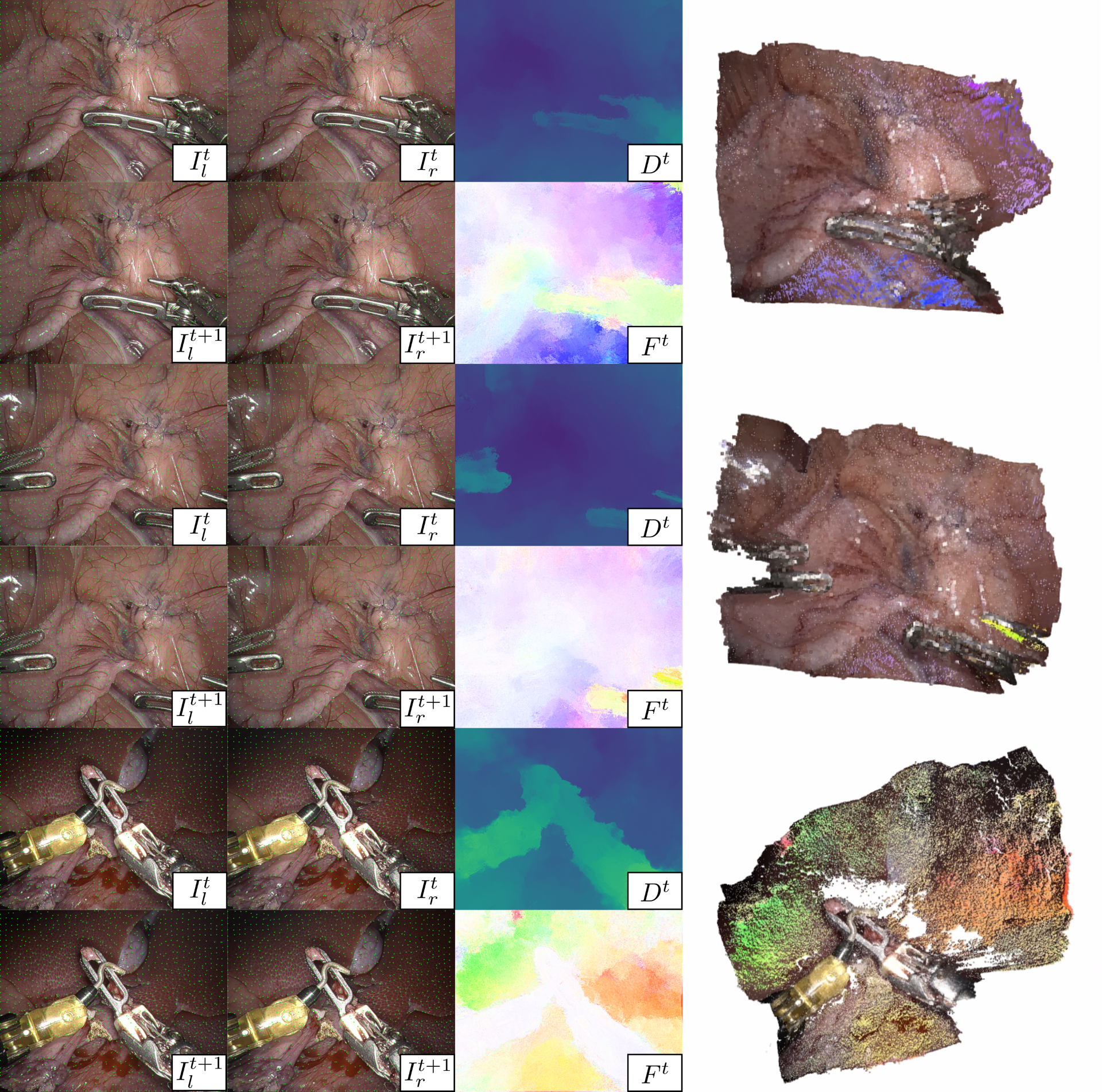}
\caption{Images with detections overlayed as green points along with stereo depth evaluations \(D^t\), and flow evaluations  \(F^t\) densely sampled for whole images. On the right are 3D point clouds of colored flow vectors \(F^t\) overlayed on tissue points backprojected using \(D^t\).}
\end{figure}
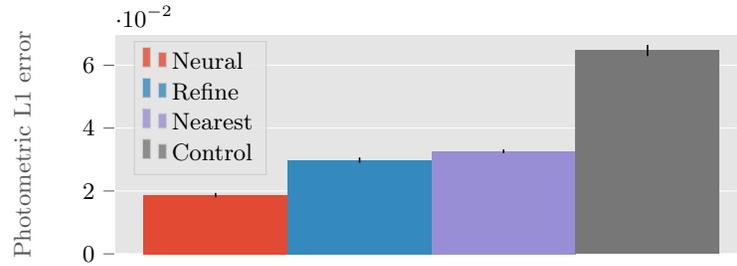
\begin{figure}[htbp]
\centering
\begin{tikzpicture}

\definecolor{chocolate2267451}{RGB}{226,74,51}
\definecolor{dimgray85}{RGB}{85,85,85}
\definecolor{gainsboro229}{RGB}{229,229,229}
\definecolor{gray119}{RGB}{119,119,119}
\definecolor{lightgray204}{RGB}{204,204,204}
\definecolor{mediumpurple152142213}{RGB}{152,142,213}
\definecolor{steelblue52138189}{RGB}{52,138,189}

\begin{axis}[
axis background/.style={fill=gainsboro229},
axis line style={white},
height=4.5cm,
legend cell align={left},
legend pos=outer north east,
legend style={
  fill opacity=0.8,
  draw opacity=1,
  text opacity=1,
  at={(0.03,0.97)},
  anchor=north west,
  draw=lightgray204,
  fill=gainsboro229
},
tick align=outside,
tick pos=left,
width=10cm,
x grid style={white},
xmajorgrids,
xmin=-0.44, xmax=0.44,
xtick=\empty,
y grid style={white},
ylabel=\textcolor{dimgray85}{Photometric L1 error},
ymajorgrids,
ymin=0, ymax=0.0697918776426285,
ytick style={color=dimgray85}
]
\draw[draw=none,fill=chocolate2267451,very thin] (axis cs:-0.4,0) rectangle (axis cs:-0.2,0.0186895541846752);
\addlegendimage{ybar,ybar legend,draw=none,fill=chocolate2267451,very thin}
\addlegendentry{Neural}

\draw[draw=none,fill=steelblue52138189,very thin] (axis cs:-0.2,0) rectangle (axis cs:-2.77555756156289e-17,0.0298158936202526);
\addlegendimage{ybar,ybar legend,draw=none,fill=steelblue52138189,very thin}
\addlegendentry{Refine}

\draw[draw=none,fill=mediumpurple152142213,very thin] (axis cs:-2.77555756156289e-17,0) rectangle (axis cs:0.2,0.0326328314840794);
\addlegendimage{ybar,ybar legend,draw=none,fill=mediumpurple152142213,very thin}
\addlegendentry{Nearest}

\draw[draw=none,fill=gray119,very thin] (axis cs:0.2,0) rectangle (axis cs:0.4,0.0646755546331406);
\addlegendimage{ybar,ybar legend,draw=none,fill=gray119,very thin}
\addlegendentry{Control}

\path [draw=black, semithick]
(axis cs:-0.3,0.0180184641157022)
--(axis cs:-0.3,0.0193606442536482);

\path [draw=black, semithick]
(axis cs:-0.1,0.0289702850172839)
--(axis cs:-0.1,0.0306615022232213);

\path [draw=black, semithick]
(axis cs:0.1,0.0320632583970767)
--(axis cs:0.1,0.033202404571082);

\path [draw=black, semithick]
(axis cs:0.3,0.0628826543685397)
--(axis cs:0.3,0.0664684548977414);

\end{axis}

\end{tikzpicture}
\caption{Average photometric reconstruction error over 512 image pairs from the porcine test split. This shows how well each phase of method performs (nearest neighbor, 3D refinement, and 3D neural interpolation (SENDD)). Control denotes estimating no motion. Error bars are the standard error of the mean.} \label{fig:photoporcine}
\end{figure}
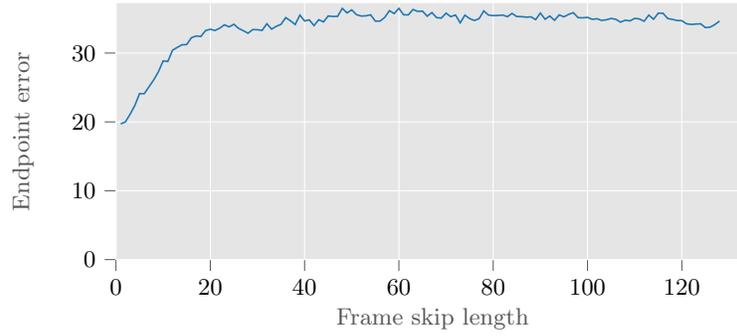
\begin{figure}[htbp]
\centering
\begin{tikzpicture}

\definecolor{dimgray85}{RGB}{85,85,85}
\definecolor{gainsboro229}{RGB}{229,229,229}
\definecolor{steelblue31119180}{RGB}{31,119,180}

\begin{axis}[
axis background/.style={fill=gainsboro229},
axis line style={white},
height=5cm,
tick align=outside,
tick pos=left,
width=10cm,
x grid style={white},
xlabel=\textcolor{dimgray85}{Frame skip length},
xmajorgrids,
xmin=0, xmax=134.35,
xtick style={color=dimgray85},
y grid style={white},
ylabel=\textcolor{dimgray85}{Endpoint error},
ymajorgrids,
ymin=0, ymax=37.3338,
ytick style={color=dimgray85}
]
\addplot [semithick, steelblue31119180]
table {%
1 19.698
2 20.003
3 21.118
4 22.385
5 24.109
6 24.099
7 25.089
8 26.091
9 27.277
10 28.85
11 28.779
12 30.421
13 30.817
14 31.199
15 31.232
16 32.23
17 32.476
18 32.401
19 33.267
20 33.456
21 33.275
22 33.608
23 34.107
24 33.798
25 34.178
26 33.589
27 33.25
28 32.875
29 33.41
30 33.386
31 33.293
32 34.262
33 33.472
34 33.881
35 34.166
36 35.148
37 34.689
38 34.142
39 35.51
40 34.658
41 34.817
42 33.984
43 34.845
44 34.519
45 35.382
46 35.313
47 35.326
48 36.487
49 35.828
50 36.273
51 35.574
52 35.374
53 35.403
54 35.553
55 34.639
56 34.641
57 35.179
58 36.159
59 35.729
60 36.494
61 35.554
62 35.544
63 36.33
64 36.046
65 36.092
66 35.347
67 35.883
68 35.138
69 35.083
70 35.796
71 35.282
72 35.53
73 34.382
74 35.501
75 35.006
76 34.732
77 34.988
78 36.117
79 35.517
80 35.432
81 35.462
82 35.522
83 35.282
84 35.755
85 35.301
86 35.289
87 35.209
88 35.263
89 34.85
90 35.809
91 34.91
92 35.381
93 34.752
94 35.54
95 35.26
96 35.617
97 35.868
98 35.142
99 35.115
100 35.181
101 34.899
102 34.985
103 34.739
104 34.833
105 35.026
106 34.902
107 34.487
108 34.772
109 34.669
110 35.039
111 34.946
112 34.619
113 35.508
114 34.907
115 35.79
116 35.779
117 35.02
118 34.888
119 34.726
120 34.703
121 34.249
122 34.157
123 34.212
124 34.246
125 33.696
126 33.76
127 34.134
128 34.639
};
\end{axis}

\end{tikzpicture}
\caption{Endpoint error on the full dataset using different frame-skips for tracking. A skip factor of two means skipping every other frame. Endpoint error decays until reaching a limit at a skip factor of around 40 frames. Different factors can be chosen for balancing efficiency and accuracy.} \label{fig:frameskip}
\end{figure}

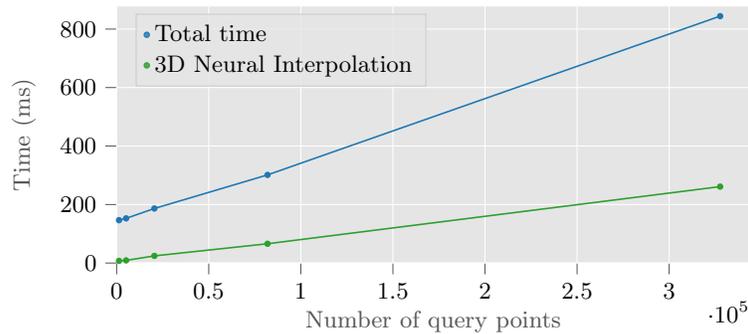
\begin{figure}[htbp]
\centering
\begin{tikzpicture}

\definecolor{dimgray85}{RGB}{85,85,85}
\definecolor{forestgreen4416044}{RGB}{44,160,44}
\definecolor{gainsboro229}{RGB}{229,229,229}
\definecolor{lightgray204}{RGB}{204,204,204}
\definecolor{steelblue31119180}{RGB}{31,119,180}

\begin{axis}[
axis background/.style={fill=gainsboro229},
axis line style={white},
height=5cm,
legend cell align={left},
legend style={
  fill opacity=0.8,
  draw opacity=1,
  text opacity=1,
  at={(0.03,0.97)},
  anchor=north west,
  draw=lightgray204,
  fill=gainsboro229
},
tick align=outside,
tick pos=left,
width=10cm,
x grid style={white},
xlabel=\textcolor{dimgray85}{Number of query points},
xmajorgrids,
xmin=0, xmax=344000,
xtick style={color=dimgray85},
y grid style={white},
ylabel=\textcolor{dimgray85}{Time (ms)},
ymajorgrids,
ymin=0, ymax=879.12902296875,
ytick style={color=dimgray85}
]
\addplot [draw=steelblue31119180, fill=steelblue31119180, mark=*, only marks,
mark size=1pt]
table{%
x  y
1280 146.4576703125
5120 152.6590671875
20480 186.4246828125
81920 301.325759375
327680 844.2399109375
};
\addlegendentry{Total time}
\addplot [draw=forestgreen4416044, fill=forestgreen4416044, mark=*, only marks,
mark size=1pt]
table{%
x  y
1280 7.2575703125
5120 8.9996890625
20480 24.338346875
81920 65.8595734375
327680 261.1905453125
};
\addlegendentry{3D Neural Interpolation}
\addplot [semithick, steelblue31119180, forget plot]
table {%
1280 146.4576703125
5120 152.6590671875
20480 186.4246828125
81920 301.325759375
327680 844.2399109375
};
\addplot [semithick, forestgreen4416044, forget plot]
table {%
1280 7.2575703125
5120 8.9996890625
20480 24.338346875
81920 65.8595734375
327680 261.1905453125
};
\end{axis}

\end{tikzpicture}
\caption{Evaluation time of SENDD as a function of the number of query points.} \label{fig:costperpoints}
\end{figure}